%% file: samu.tex
\documentclass[a4paper]{article}
\pdfoutput=1

\usepackage{dblfloatfix}
\usepackage{caption}  

\usepackage{graphicx}
\usepackage[numbers]{natbib}

\usepackage{tikz}
\usetikzlibrary{shapes,arrows,calc,positioning}

\usepackage{amssymb}
\usepackage{amsmath}

\usepackage{algorithm}
\usepackage{algorithmicx}
\usepackage{algpseudocode}

\usepackage{subfigure}
\usepackage{paralist}

\usepackage{listings}
\usepackage{xcolor}

\usepackage[export]{adjustbox}

\usepackage{url}

\newcommand{\keywords}[1]{\par\addvspace\baselineskip\noindent\textbf{Keywords: }\textit{#1}.}

\begin{document}
\title{A disembodied developmental robotic agent called Samu B\'atfai}

\author{Norbert~B\'atfai,
\thanks{N. B\'atfai is with the Department of Information Technology, University of Debrecen, H-4010 Debrecen PO Box 12, Hungary, e-mail: batfai.norbert@inf.unideb.hu.}}

\maketitle

\begin{abstract}
The agent program, called Samu, is an experiment to build a disembodied 
DevRob (Developmental Robotics) chatter bot that can 
talk in a natural language like humans do.
One of the main design feature is that Samu can be interacted with using only a character terminal. 
This is important not only for practical aspects of Turing test or Loebner prize, but also for  
the study of basic principles of Developmental Robotics. Our purpose is to create a rapid prototype of 
Q-learning with neural network approximators for Samu. We sketch out the early stages of the development process of this prototype, where Samu's task is to predict the next sentence of tales or conversations. 
The basic objective of this paper is to reach the same results using reinforcement learning with general function approximators that can be achieved by using the classical Q lookup table on small input samples. 
The paper is closed by an experiment that shows a significant improvement in Samu's learning when using LZW tree to narrow the number of possible Q-actions.
\end{abstract}


\section{Introduction}
 
At first glance, it seems impossible to develop a natural language chatter bot that 
can understand 
conversations as humans do, 
and that is only based on the verbal content of the speech because human communication is only 7 percent verbal \cite[pp. 26]{DevRobSurvey}. 
Nevertheless we have started to develop such a system, called Samu. Or, to be more precise, it is called Samu B\'atfai due to it will be taught primary in a family circle 
where the author's three children, ages 7 and 9 years old will also be partially its caregivers.

One of the main design feature of agent program Samu is that we can communicate with it through only one written channel. For example, a consequence of this is that we cannot create associations between  a visual channel and an auditory channel. We must work only from the verbal content, but this constraint is what allows us to be able to purely focus on the investigation of basic principles of Developmental Robotics \cite{DevRobPrinc}.
And there is a further practical reason why Samu has only limited input capability. The character channel will be controlled and monitored very well by caregivers. It is not of great importance at this moment but it will be much more important in the future.   
Because if we think about it, we are not to permit either our little children or our little DevRob agents to use internet without supervision.

This paper is organized as follows. First, it introduces the architecture of Samu. Then it explains how Samu is linked to Developmental Robotics in a conceptual level. Finally we evaluate our results supplemented with further conceptual elements.

\section{Samu}

The verification principle  \cite{DevRobPrinc} is the most important principle of Developmental Robotics. 
At the beginning of the developmental process, program of Samu will not possess the ability to verify the knowledge. Mostly this is because, like young children of age 2-4 years old, young Samu accepts what its caregivers say without reservation. That is why family caregivers have their own dedicated communication channels on standard input/output. Using these channels, Samu will be able to distinguish its family caregivers, as it is shown in the next lines

\lstset { %
    language=C++,
basicstyle=\ttfamily\footnotesize
}
\begin{lstlisting}
Norbi@Caregiver> ___next caregiver
Nandi@Caregiver> I am Nandi
\end{lstlisting}

where the message started with the prefix \_\_\_ (three underscores) are not sent directly to Samu.
Talking on the internet with other conversationalists and especially with robots that are similar to Samu will be available at a later development stage. In both of these stages, the verification will be able to be carried out only on the basis of Samu's accumulated knowledge and experience.
It seems obvious that the knowledge acquisition of a developmental robotic agent can be programmed in a natural way by using some reinforcement learning method. The general architecture of a system of this kind is shown in the following section.

\subsection{An idealized architecture}

The paper \cite{NatureQ} gave the idea for the idealized architecture of Samu that is shown in Fig. \ref{fig_idealarc} where
Samu reads the sentences from a children's tale or from a conversation. Then it tries to reconstruct the meaning of the read sentences  in the form of a  2D simulation game generated from the sentences. Like in paper \cite{NatureQ}, the screenshots of the generated game will be processed by a Q-learning network, but whereas there the scores are part of the game, here they are based on the comparison of the reality of the reconstructed game and the reality of the input sentences. Theoretically, it would be a hybrid architecture that combines a model-free (Q-learning) and a knowledge-based (generating the 2D programs) approaches. To clarify matters, this architecture would use a model-free method to develop knowledge-based models. 
But because worth of every model is what we can implement from it, we will strongly simplify it in the next sections.

\subsection{Simplified architectures}

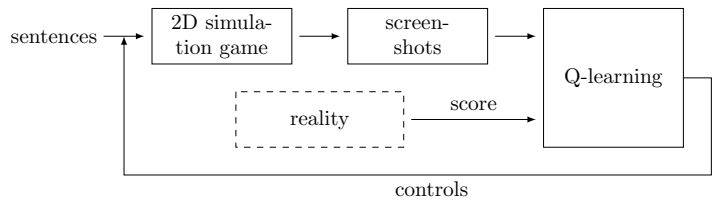
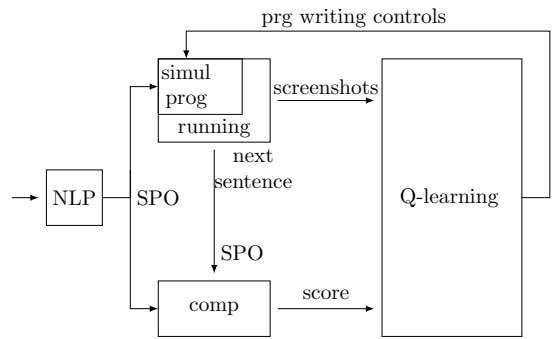
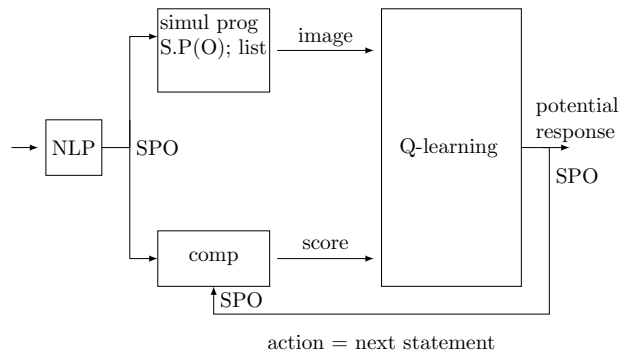
\begin{figure*}[h!]
\centering
\subfigure[An idealized architecture for planning Samu. As far as it will be possible, the comparison of the reality of the reconstructed game and the reality of the input sentences will have to be automated.]{
\scalebox{.8}{\input{samuarc1}}
\label{fig_idealarc}}\\
\subfigure[This modified architecture enables fully automated comparison between the reality and the simulation. The score represents the positive reward if the previous predicted (simulated) sentence matches the actual input string. And a negative reward is given if the predicted (simulated) sentence of the previous step differs from the actual input string.]{
\scalebox{.8}{\input{samuarc2}}
\label{fig_arc2}}
\subfigure[This is a simplified version of the previous architecture where the next sentence is determined by the action of Q-learning. And notice that the simulation program is not being run. Instead, the state of the Q-learning is only the "source code" itself. It is a very simplified simulation program that consists only of statements that have the following form: Subject.Predicate(Object).]{
\scalebox{.8}{\input{samuarc3}}
\label{fig_arc3}}
\caption{This figure shows some architectural sketches that build on each other to help refining the initial concept of Samu.\label{fig_arc123}}
\end{figure*}

Starting from the idealized architecture shown in Fig. \ref{fig_idealarc} we base the comparison of 
the reality  and the simulated one derived from the input sentences 
on the principle of consciousness oriented programming 
(the agents must predict the future and the conscious ones can really foresee and predict it \cite{cop}). It is shown in Fig. \ref{fig_arc2} where the simulation program's task is to predict the next input sentence. This modified architecture works with SPO triplets (Subject-Predicate-Object, \cite{spo}) rather than the raw sentences being read. 
Fig. \ref{fig_arc3} shows a further simplified architecture.  In this case, we do not have to run the simulation program to predict the next input sentence, instead the prediction is done by the Q-learning component itself.
The architectural components of Fig. \ref{fig_arc123} are discussed in detail in the next points.

\subsection{The first rapid prototype}

We use a hybrid architecture of the latter two sketches in Fig. \ref{fig_arc123} for implementing the first rapid prototype of Samu. 
In the following, this architecture is referred to as \textit{Samu's cognitive engine}. 

\subsubsection{NLP}

As we mentioned earlier, SPO triplets \cite{spo} are used instead of sentences. For example, if Samu reads the sentence \textit{"The little mouse has become a big lion."} the NLP (Natural Language Processing) component should produce the following triplets:  
  \begin{inparaenum}[i)]
    \item \textit{"(mouse, become, lion)"}
    \item \textit{"(mouse, is, little)"}
    \item \textit{"(lion, is, big)".}
  \end{inparaenum}
Link Grammar \cite{LinkGrammar} is used to obtain SPO triplets but automatically identifying these is a complex data mining problem. At this moment we are not focusing on this matter. We are going to partially implement some algorithms of \cite{spo2} but for this paper, we use only a very simple algorithm shown in code snippet in Fig. \ref{fig_nplhpp}.

\begin{figure*}[h!]
\lstset { %
    language=C++,
basicstyle=\ttfamily\footnotesize
}
\begin{lstlisting}
for ( int k {0}; k<linkage_get_num_links ( linkage ); ++k ) {
    const char* c = linkage_get_link_label ( linkage, k );
    if ( *c == 'S' ) {
        triplet.p = linkage_get_word ( linkage, k );
        alter_p = words[linkage_get_link_rword ( linkage, k )];
        triplet.s = words[linkage_get_link_lword ( linkage, k )];
    }
    if ( *c == 'O' ) {
        triplet.o = words[linkage_get_link_rword ( linkage, k )];
        if ( triplet.p == words[linkage_get_link_lword ( linkage, k )] ) {
            triplets.push_back ( triplet );
        } else if ( alter_p == words[linkage_get_link_lword ( linkage, k )] ) {
            triplet.p = alter_p;
            triplets.push_back ( triplet );
        }
    }
}
\end{lstlisting}
\caption{This is a simplified code snippet that is iterated over all linkages of the parsed sentence returned by the Link Grammar \cite{LinkGrammar}. The '\texttt{words}' is a vector of strings that contains all words of a given linkage. The complete source code in which this snippet is incorporated can be found in the file \texttt{nlp.cpp}. This transitional solution can handle the simplest sentences like 
"A rare black squirrel has become a regular visitor to a suburban garden"
, "This is a car"
, "This car is mine"
, "I have a little car"
, "The sky is blue"
, "The little brown bear has eaten all of the honey"
, "I love Samu".
It may be noted that the first one is the example sentence used in the paper \cite{spo}.
 \label{fig_nplhpp}}
\end{figure*}

\subsubsection{Visual imagery}

\begin{figure}[!h]
\centering
\includegraphics[scale=.5, frame]{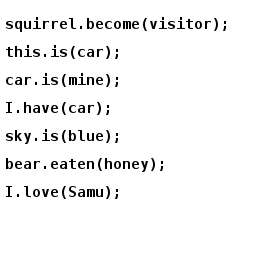}
\caption{This is a sample mental image (256x256 of size) that contains sentences shown in the caption of Fig. \ref{fig_nplhpp}.}
\label{fig_vi}
\end{figure}

To describe the simulation programs we use a very simplified object oriented style language, where statements have the following form: 
\[
\texttt{Subject.Predicate(Object);} 
\]
where each identified triplet corresponds to a statement such as this.
A simulation program is a statement list that only contains a given number or less statements. 
In addition, we  write it into an image because it would be difficult to handle the statement list as a textual list of statements as it is shown in Fig. \ref{fig_arc3} where the input of the Q-learning is already an image. And finally, it may be noted that the sequence of such "mental" images can be interpreted as visual imagination of Samu. Fig. \ref{fig_vi} shows a sample mental image.

\subsubsection{Q-learning}

Sigmoidal multilayer perceptron (MLP) neural networks are used to approximate the Q-function.
In accordance with the QCON model \cite{qcon1}, \cite{qcon2} each possible action has an unique neural network to approximate the Q-value associated with the given state. 
Each MLP contains three layers with 65536 input units, 80 hidden layer units and an output neuron which gives the approximation of the Q-value of the state-action pair. The states are "mental" images introduced in the previous paragraph. 
Several examples can be found in the literature of using neural networks to approximate the Q-function, for example, see the book \cite{GosaviBook} or the tutorial \cite{GosaviTutorial} by Gosavi. 
Alg. \ref{alg-sq} shows the used algorithm to perform the Q-learning. This pseudo-code is a simplified and generalized version of the most precise description of the used algorithm that can be found in the source file \texttt{ql.hpp} at \url{https://github.com/nbatfai/samu/blob/master/ql.hpp}. The regular part of the algorithm is taken from the book \cite[pp. 844]{RussellNorvig}.

\begin{algorithm}[h!]
\small
\caption{Q-learning-with-NN-approximation ($s'$, $t'$) \textbf{returns} a triplet (an action)\label{alg-sq}}
\begin{algorithmic}[1]
\Require $s'$, $t'$\Comment{$s'$ is the current state and $t'$ is the current triplet.}
	\Statex $r', a', p, nn, q$ \Comment{Local variables: $r'$ is the current reward, $a'$ is the current action and $p$ is a perceptron,  $nn, q \in \mathbb{R}$.}
	\Statex $s, r=-\infty, a, N_{sa}, Prcps_{a}$\Comment{Persistent variables: $s$ is the previous state, $r$ is the previous reward, $a$ is the previous action, $N_{sa}$ is the state-action frequency table and $Prcps_{a}$ is the array of the perceptrons.}
\Ensure $a'$\Comment{$a'$ is an action, the predicted next triplet.}
\Procedure{$\operatorname{SamuQNN}$}{$s'$, $t'$}
\State $r' = compare(t', a)$ 
\State $a' = t'$ 
  \If{$r > -\infty$}
	\State Increment $N_{sa}[s,a]$
	\State $nn = Prcps_{a}[a](s)$
	\State $q = nn +\alpha(N_{sa}[s,a]) (r' + \gamma\ max_{p} Prcps_{a}[p](s') - nn)$
	\State Back-propagation-learning($Prcps_{a}[a], s, q, nn$) \Comment{where $q-nn$ is the error of the MLP $Prcps_{a}[a]$ for input $s$.}
	\State $a' = argmax_{p} f(Prcps_{a}[p](s'), N_{sa}[s',p])$
\EndIf
\State $s = s'$
\State $r = r'$
\State $a = a'$
  
\State \Return{$a'$}
\EndProcedure
  \end{algorithmic}
\end{algorithm}

\subsection{Conceptual background of Samu}

In order to outline the conceptual background we should begin with the explanation of the usage of the notion of disembodiment.
It is simply an indication that Samu project is \textit{100 percent pure software}. We can say that Samu is a robot without a body. Certainly, this does not mean a return to GOFAI.
All the more so, because we are going to follow the recommendations given as replies to the critics of Dreyfus and Dreyfus \cite[pp. 1025]{RussellNorvig}. 

It is certainly arguable whether the principle of embodiment is met or not. We believe that we are not confronted with this principle as it will also be shown in the following.

Returning to the basic principles formulated by \cite{DevRobPrinc}, there are two different use cases in respect of the validation processes of \textit{Samu's cognitive engine}. If we observe how Samu operates we can easily distinguish a short-term verification from a long-term one. It is clear that the short-term verification of Samu's knowledge is trivial because it is the prediction itself. This type of verification is a quintessence of the whole system. But the long-term verification is a much harder issue since this is only a naive intuitive expectation. The long-term verification should give the answer to the question: How can we extend the short-term verification for longer time periods. In this sense the long-term verification means the understanding the conversations and read texts. It is an open question how much the distance between these two kinds of verifying the knowledge. Based on our experiences we have used a quantity called \textit{bogo-relevance} that may be positioned somewhere in the middle of these two extremes of verification. Using the notations of Alg. \ref{alg-sq} the computation of the \textit{bogo-relevance} of a prediction $q$ in the state $s$ can be rewritten as follows:
\begin{align*}
\operatorname{bogo\_relevance}(s, q) &= \\
=&\frac{Prcps_{a}[q](s) - \overline{Prcps_{a}}}{\max\limits_{p} Prcps_{a}[p](s) - \min\limits_{p} Prcps_{a}[p](s)}
\end{align*}
where the $\overline{Prcps_{a}}$ is the mean of the computed values of the perceptrons that are applied to the same given input state $s$, that is
\[
\overline{Prcps_{a}} = \frac{\sum\limits_{p \in  Prcps_{a}}{Prcps_{a}[p](s)} }{\vert Prcps_{a}\vert}
\]
here $\vert Prcps_{a}\vert$ denotes the number of all perceptrons in the structure $Prcps_{a}$. The program prints the values $100 * \operatorname{bogo\_relevance}(s, q)$ that can be seen in Fig. \ref{fig_tui}.

From a spiritual point of view, Samu's soul consisting of the weights of Samu's ANNs.
Because they are only simple numbers (and not quantum qubits, for example), Samu's soul can be saved to a file. 
This file is called \texttt{samu.soul.txt} and it is created automatically when Samu's program receives a given (SIGINT, SIGTERM, SIGKILL or SIGHUP) signal. But since the soul cannot live without body, Samu's body is the UNIX process that runs the source code of Samu. The actions of Samu are his statements and replies that are formed by the close interaction with the external world. In the light of this we think that, we are not confronted with the spirit of the principle of embodiment \cite{DevRobPrinc}.

The fulfillment of the principle of subjectivity \cite{DevRobPrinc} is trivially met since it follows from using human caregivers to train Samu. Accordingly, the a priori subjectivity of caregivers implies the subjectivity of their robots. 

It is easy to see that the criteria of the principle of grounding \cite{DevRobPrinc} have also been fulfilled during the short-term verification. Because the set of possible outcomes \{good prediction, bad prediction\} may be assigned to each action, since each action is a predicted sentence in our model.

Finally, it may be noted that in the sense of the terminology introduced in \cite{cop}, Samu may become a conscious or even an intuitive computer program since it will be able to predict the future better than a random guess. 

\section{Results}

The results and  evaluations are divided into three experiments. In the first experiment, we verify Samu's implementation of Q-learning by comparison with using the classical Q lookup table. Here we have only used a small number of sentences. 
The second experiment extends the investigation to a larger sample. 
Finally, the third experiment has already focused only on the learning of the larger corpus where using the Liv-Zempel-Welch (LZW) \cite{lzw} dictionary tree to narrow the scope of selecting Q-actions seems promising for accelerating Samu's learning.

\subsection{Experiment 1:  "SAMU A"}

\begin{figure}[h!]
\centering
\includegraphics[scale=.5]{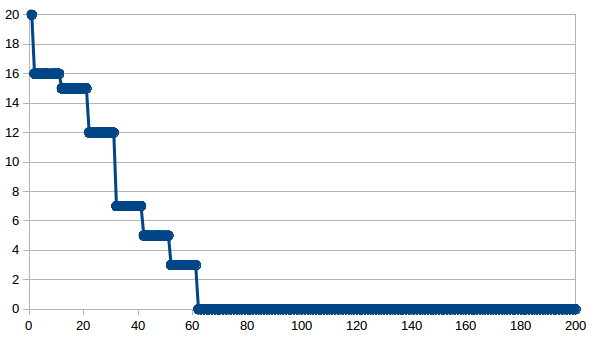}
\caption{The horizontal axis shows the number of trials using Q lookup table.
Trials consist of reading the 7 given sentences. 
The vertical axis shows the prediction error of the next sentences.\label{fig_q}}
\end{figure}

\begin{figure}[h!]
\centering
\includegraphics[scale=.5]{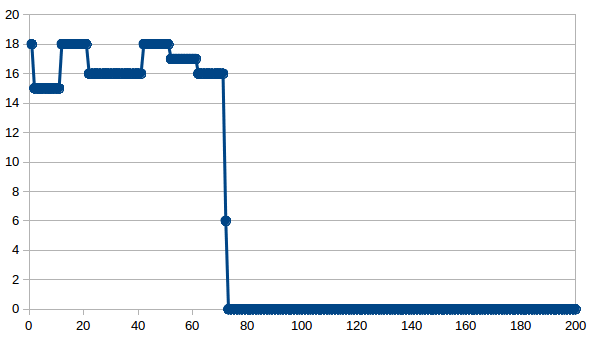}
\caption{The horizontal axis shows the number of trials using general function approximators. The vertical axis is the same as in the case of Fig. \ref{fig_q}.\label{fig_nn}}
\end{figure}

Our main purpose in this paper is to equip Samu with the ability to carry out reinforcement learning with general function approximators. To validate this we use the following experiment. 
The sentences shown in the caption of Fig. \ref{fig_nplhpp} are considered as a short story and Samu is being trained continuously with this during the experiment. Samu reads the sentences of the story and predicts the next story sentence. 
A third of a point is given if a triplet member is predicted correctly. The reward is based on the comparison of triplets and is computed by the following equation
\lstset { %
    language=C++,
basicstyle=\ttfamily\footnotesize
}
\begin{lstlisting}
    double reward =
      3.0 * triplet.cmp ( prev_action ) - 1.5;
\end{lstlisting}
therefore 10.5 points are given if all predictions are correct and conversely -10.5 points are given if all predictions are wrong. We implement both the simple Q lookup table and the general function approximators based solutions in order to compare them. 
The learning curves for these two experiments can be seen on Fig. \ref{fig_q} and Fig. \ref{fig_nn}. For example, the first one shows that Samu can perfectly  memorize the text (consisting of the seven test sentences) after he read it roughly sixty times.
It can be seen well on Fig. \ref{fig_nn} that the general function approximators based solution can also learn the true predictions.
So we may assume that our neural network based approximation works well. 
Therefore, in the following we are not going to use the Q lookup table either in this paper or in the source codes. (Lookup table-based code snippets can still be found in the source files. These have been wrapped in conditional compilation directives but they are deprecated in the new versions of Samu, see for example \cite{isaacrepo}, \cite{jacobrepo}, \cite{judahrepo}.)

In order to allow the evaluation and verification of these results, we create a GitHub repository for the source of this study. 
It is called \texttt{samu} and available at \url{https://github.com/nbatfai/samu}. 
The two use cases of Samu shown in Fig. \ref{fig_q} and Fig. \ref{fig_nn} are presented in a YouTube video at the following link \url{https://youtu.be/qAhxVQk-dvY}.
It is our hope that Samu will be the ancestor of DevRob chatter bots that will be able to chat in natural language like humans do. 
However, it should be noted that in order to use Samu as a chat program we need to try another fork of the project Samu \cite{samurepo}.
The first official forks, called Isaac \cite{isaacrepo}, Jacob \cite{jacobrepo}, Judah \cite{judahrepo}, Perez \cite{perezrepo}, Hezron \cite{hezronrepo}, Ram \cite{ramrepo} and Amminadab \cite{amminadabrepo} are introduced in next section.  

\subsubsection{Samu's family tree}

Samu and his descendants (at this moment Isaac, Jacob, Judah, Perez and Hezron) are open-source programs released under the GNU GPL v3 or later. Samu is a generic name of these projects that are intended to investigate deep reinforcement learning methods to facilitate the development of a chat robot that will be able to talk in natural language like a human does. At this moment they are case studies of using Q learning with neural networks for predicting the next sentence of a conversation. They can only be run on Unix-type (such as GNU/Linux) environments.

\begin{figure}[h!]
\centering
\includegraphics[scale=.47]{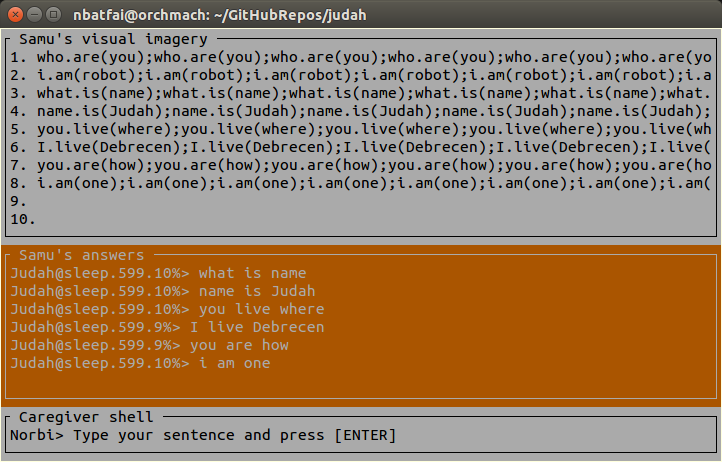}
\caption{A screenshot of Judah's TUI. The top window contains the contents of Samu's visual imagery. The mid window displays Samu's responses.
For example, the prompt \texttt{Judah@sleep.599.9\%} contains the name of the robot, an at sign (@),  the state of the robot,  the number of the  known triplets and a \textit{bogo-relevance} value. The bottom window is the text input field where the caregiver can enter text and commands for Samu.}
\label{fig_tui}
\end{figure}

\begin{inparaenum}[\itshape a\upshape)]
\item Samu \cite{samurepo} is the initial project to start creating a DevRob chatbot agent. This project is presented in detail in this paper.
Samu's development is frozen to allow the evaluation and verification of the results shown in Fig. \ref{fig_q} and Fig. \ref{fig_nn}.
Therefore this project is continued in the project entitled Isaac.
\item The project Isaac \cite{isaacrepo} has been forked from Samu. Isaac takes a step towards the deep Q learning, for example, Isaac's perceptrons are accelerated by using CUDA and OpenMP.
\item The project Jacob \cite{jacobrepo} has been forked from Isaac. Jacob replaces Isaac's picture-based visual imagination with a character-based one. 
\item The project Judah \cite{judahrepo} has been forked from Jacob. Judah equips Jacob with a text-based user interface (TUI) that can be seen in Fig. \ref{fig_tui}. In addition, Judah can also be run on both remote computers as well as on supercomputers.
\item The project Perez \cite{perezrepo} has been forked from Judah. Using this, we can make experiments to see the behaviour of Samu on different parameter settings of the incremental learning. 
\item The project Hezron \cite{hezronrepo} has been forked from Perez. In this project we are going to compare the arrangements of SPO triplets in the visual imagery. For example, a pyramid-shaped arrangement can be seen in a YouTube video at \url{https://youtu.be/zjoINedftPY} or a fully justified one is shown in Fig. \ref{fig_tui}. 
\item The project Ram \cite{ramrepo} has been forked from Hezron. Ram allows to experiment with different deep Q learning algorithms to express feelings when he is talking.
\item The project Amminadab \cite{amminadabrepo} has been forked from Ram. Amminadab uses the LZW tree to narrow the scope of selecting Q-actions. 

\end{inparaenum}

\subsection{Experiment 2:  "JUDAH A"}

\begin{figure}[h!]
\centering
\includegraphics[scale=.5]{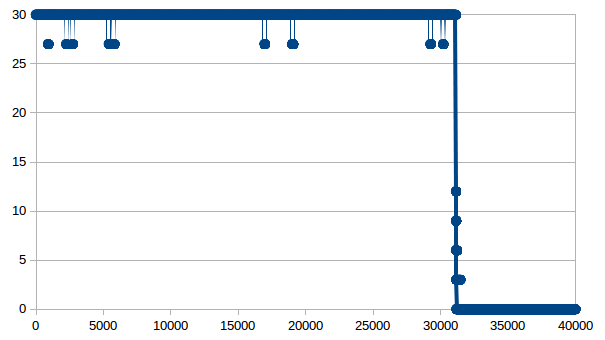}
\caption{This graph is the same as the previous one except that before the measurement, Samu has been trained with the bigger corpus for several thousand times. The axes are the same as in the previous figure.\label{fig_expa2}}
\end{figure}

\paragraph{Religious robot children}
As we have already mentioned earlier, in this paper we focus on the shaping of the neural infrastructure of Samu rather than on the triplet extraction. But certainly we need corpora for training and testing Samu. Evidently that we used only a meaningless sequence of sentences in the previous example of the seven test sentences.  It was enough to verify the implementation of the used algorithms but  it is not suitable for teaching a chat robot. We need real corpora and it is equally important to use sophisticated triplet extraction algorithms that can preprocess these corpora.
In addition, from the point of view of Developmental Robotics, the essential constraint of the selection of the training corpora is that these may only contain a limited number of distinct words. An example of such corpus may be the Basic English Bible (BBE, Bible in Basic English) \cite{bbe1} that contains 850 words \cite{bbe2}. 
We have tried to use the Gospel of Matthew.
But it should be noted that the initial simple algorithm shown in Fig. \ref{fig_nplhpp} can extract only roughly 700 triplets from it. 
This is due to the following two reasons. 
\begin{inparaenum}[\itshape i\upshape)]
\item The used algorithm is elementary in the sense that it tries to apply only a very simple rule to extract triplets as exactly shown in  Fig. \ref{fig_nplhpp}
\item In the raw text source of BBE, the '.', ':' and ';' letters were replaced by a newline character and the numbers were deleted when we applied the triplet extraction mechanically to each line. (And for example, there were lines that contains only one word. The results of this preprocessing can be found in the files called \texttt{bbe} and \texttt{bbe.triplets} that are available at the author's university page \url{http://shrek.unideb.hu/~nbatfai/}.)
\end{inparaenum}
It follows that this corpus of the 700 SPOs still has not been regarded as a text that can hold meaning for human readers. 
In a second experiment, we use this corpus and another small one consisting of the sentences shown in Fig. \ref{fig_intro}.
\begin{figure}[h!]
\lstset { %
    language=C++,
basicstyle=\ttfamily\footnotesize
}
\begin{lstlisting}
    {
      "introduce myself",
      {
        "Who are you",
        "I am a robot",
        "What is your name",
        "My name is Judah",
        "Where do you live",
        "I live in Debrecen",
        "How old are you",
        "I am one year old",
        "Where were you born"
        "I was born is Debrecen"
      }
    }
\end{lstlisting}
\caption{A small training corpus that has already held meaning for human readers.
 \label{fig_intro}}
\end{figure}
Because of the larger number of all sentences of these two used corpora we give a more sharp comparison for triplets and the rewarding policy. 
The reward is changed by the following equation
\lstset { %
    language=C++,
basicstyle=\ttfamily\footnotesize
}
\begin{lstlisting}
    double reward =
      ( triplet == prev_action ) ? 1.0 : -2.0;
\end{lstlisting}
therefore in the case of the introductory text 10 points are given if all "responses" are correct and conversely -20 points are given if all "answers" are incorrect. The size of the input and hidden layers of MLPs is reduced due to we use a character console-based visual imagination (that was introduced by the project Jacob \cite{jacobrepo}).  
The neural networks corresponded to the triplets are equipped with 800 input units and 32 hidden layer units in this experiment.
Fig. \ref{fig_expa1}, Fig. \ref{fig_expa2} and Fig. \ref{fig_expa1-25} show the learning curve for teaching the smaller (the introductory) text contained in Fig. \ref{fig_intro}.
\begin{figure}[h!]
\centering
\includegraphics[scale=.5]{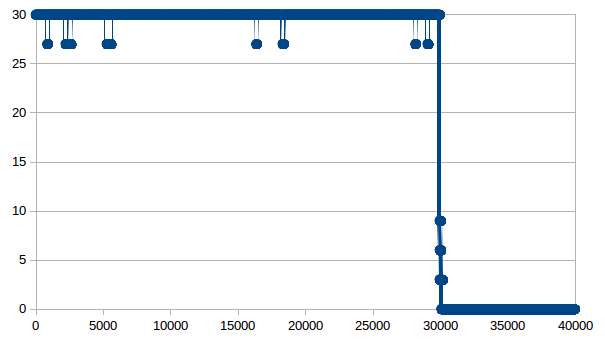}
\caption{The learning curve for the introductory text contained in Fig. \ref{fig_intro}. The axes are the same as in Fig. \ref{fig_q} and \ref{fig_nn}. This graph can be verified and reproduced by using Judah's git repository \cite{judahrepo} tagged as \texttt{v0.0.1.exp.a1}. The execution takes a couple of hours on an ordinary PC.  A detailed description of using the program can be found in the Judah project's wiki at \texttt{https://github.com/nbatfai/judah/wiki}.\label{fig_expa1}}
\end{figure}
\begin{figure}[h!]
\centering
\includegraphics[scale=.5]{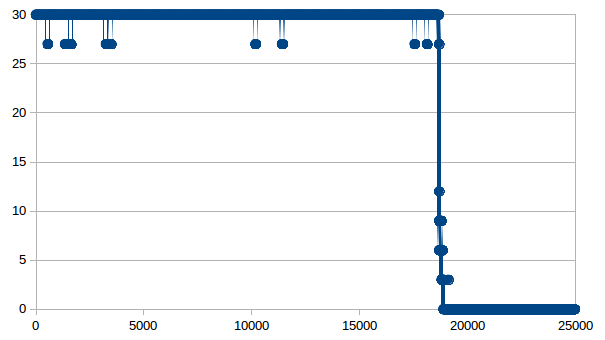}
\caption{In this figure, the learning curve of Fig. \ref{fig_expa1} is recomputed with the modified number of tries for triplets in the exploration function. There is a YouTube video at \texttt{https://youtu.be/ONgfQDYlqvo} that shows how to use Judah to reproduce this figure.\label{fig_expa1-25}}
\end{figure}

In connection with using sacred texts for training Samu's neural networks, the question arises whether religious training should be given to robots? Samu may transform this question an theoretical-ethical one into a practical one. In a more pragmatic sense, it would be an interesting linguistic and theological challenge to attempt to make a new translation of a Gospel into SPO triplets.

\paragraph{Family circle robotics}
Another possibility to learn Samu's neural networks is to use the conversational interactions of CHILDES (Child Language Data Exchange System) \cite{childes}. But we believe that the social interactions with family caregivers will become the mainstream in teaching Samu. For this reason, Samu saves all conversations that take place with the caregivers as training files.

\subsubsection{Social interaction with each other}
Perhaps the most important possibility is that when the Samu-type DevRob agents share their knowledge among each other.
At this moment the development of the communication protocol is in progress.

\subsection{Experiment 3:  "JUDAH and PEREZ B"}

\begin{figure}
\centering
\includegraphics[scale=.5]{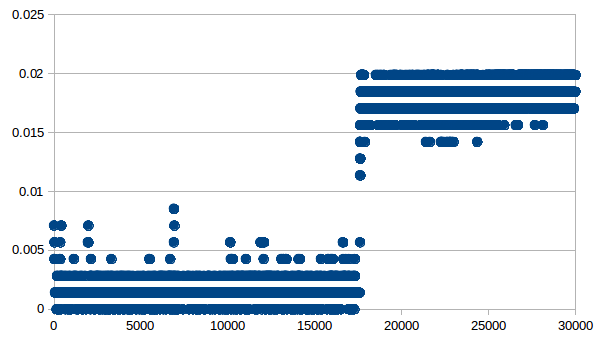}
\caption{In this figure we have not used the reward-based scale that was used in the previous ones. Now the vertical axis simply shows the quotient of number of good predictions and the sum of the good and the bad ones. For example, the value .02 means roughly 14 good and 690 bad predictions. This experiment has been done with a variant of the Judah version tagged by \texttt{v0.0.2.exp.b1} (the name of the BBE corpus must be specified in line 236 (see at {\tt https://github.com/nbatfai/ju\-dah/blob/mas\-ter/main.cpp\#L236}). In sum, Samu has not learnt the larger corpus yet in this experiment).\label{fig_expB1}}
\end{figure}

\begin{figure}
\centering
\includegraphics[scale=.5]{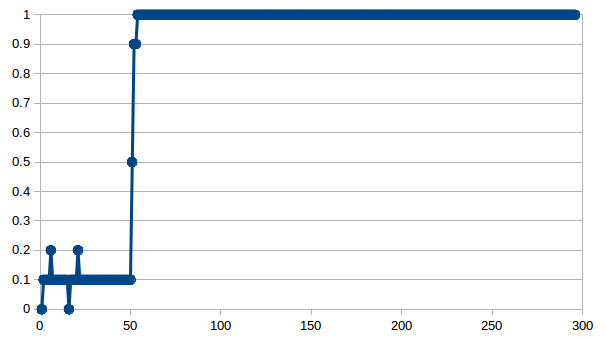}
\caption{This graph shows that Samu succeeded in learning the smaller corpus. The source code for this experiment can be found on the Judah repository \cite{judahrepo} tagged by \texttt{v0.0.2.exp.b1} (see also at \texttt{https://github.com/nbatfai/ju\-dah/re\-leas\-es/\-tag/\-v0.0.2.exp.b1}).\label{fig_expB2}}
\end{figure}

\begin{figure}
\centering
\includegraphics[scale=.5]{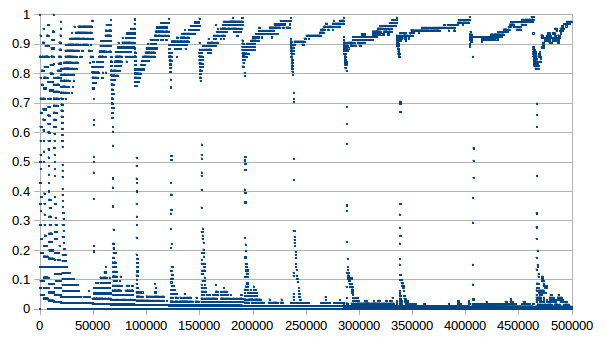}
\caption{Samu managed to learn the larger corpus now. 
To be more precise he has learnt the first 126 (18*7) sentences out of roughly 700 (during 18 incremental steps).
The source code for this experiment can be found on the Perez repository \cite{perezrepo} tagged by \texttt{v0.0.1.exp.b1} (see also at \texttt{https://github.com/nbatfai/perez/\-releases/\-tag/\-v0.0.1.exp.b1}). For achieving the same learning curve like this the cache file \texttt{bbe.triplets} needs to exist in the directory from where Samu was executed.\label{fig_expB3}}
\end{figure}

\begin{figure}
\centering
\includegraphics[scale=.5]{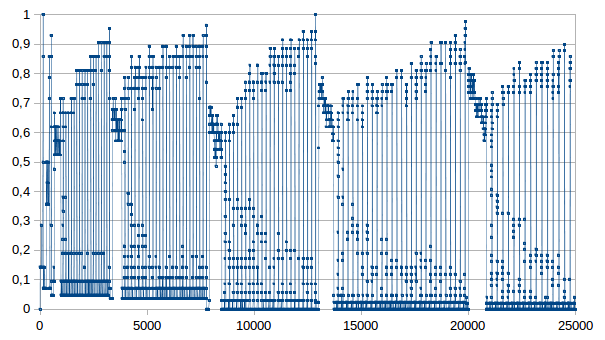}
\caption{This figure shows only the first 25000 points of the previous one. It is interesting to note that the success of the learning of the incremental training corpora (when the quotient in question is greater than roughly 0.95) is followed by a short stable period in which there is no significant forgetting.\label{fig_expB31}}
\end{figure}

This experiment is the same as the previous one, but here Samu learns the larger corpus. It should be noticed that perfectly memorizing a verbatim text of 700 sentences is really difficult or nearly impossible for most humans, as well. Accordingly, the algorithms developed in Judah have not performed well. Fig. \ref{fig_expB1} shows that Samu (to be more precise Judah) learns very slowly and it may even be that he does not learn. We must try to accelerate the learning. A promising acceleration mechanism is tested with the smaller corpus in Fig. \ref{fig_expB2}. 
We have modified the usage of the exploration function. In addition we use a trick of incremental learning \cite{incremental}, \cite{forgetting}.
Using a trivial version of this, Samu already can learn the larger corpus. It is shown in Fig. \ref{fig_expB3} and Fig. \ref{fig_expB31}.
In the beginning, we taught only the first seven sentences of the larger corpus as a training text.
After Samu had already learnt it, we added the next seven sentences of the larger corpus to the training file. This incremental procedure was repeated continuously during the experiment.
However it is clear that it will not be enough if the number of triplets continues to increase. 
For example, during the experiment, it was well felt that the phenomenon of catastrophic forgetting \cite{forgetting} has slowed the speed of Samu's learning.
This is also reinforced by the fact that the incremental learning was implemented in the \texttt{main.cpp} and not in a class of Samu. From a point of view of a programmer, it means that the incremental approach is not yet really part of Samu.
We think there may be another possibility to improve the performance of Samu's neural networks.
In our near future work, we are going to carry out experiments to see whether it is possible to increase the speed of learning by improving Samu's visual imagery. Experimenting with several (such as pyramid or justified) arrangements of triplet statements in the visual imagery does not bring improvements directly. A more interesting possibility would be to add new internal information to the mental images. These information might be interpreted as feelings, for more details  in this direction see the project Ram \cite{ramrepo}.

\subsection{Experiment 3:  "AMMINADAB B"}

\begin{figure}
\centering
\includegraphics[scale=.5]{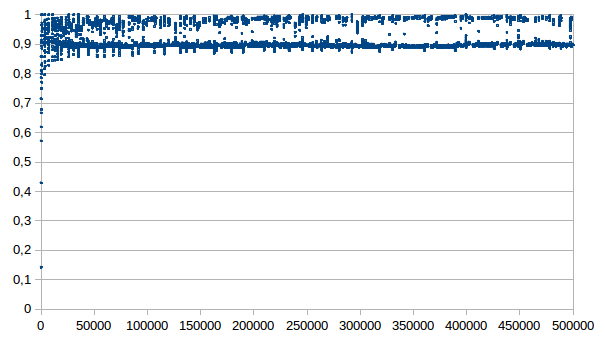}
\caption{This figure repeats the experiment shown in Fig. \ref{fig_expB3} by using Amminadab project  \cite{amminadabrepo}. It shows very well that Amminadab can significantly improve Samu's learning curve. He has already learnt the first 392 (56*7) sentences out of 704 (during 56 incremental steps).
\label{fig_lzw1}}
\end{figure}

\begin{figure}
\centering
\includegraphics[scale=.5]{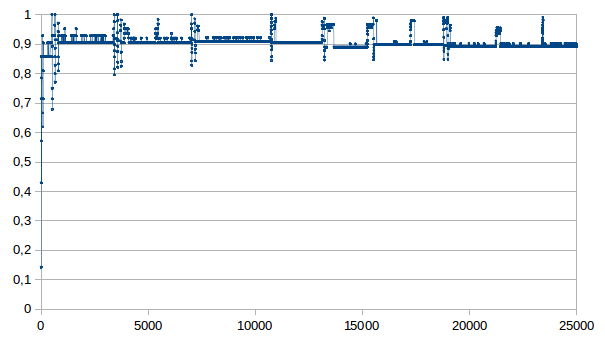}
\caption{Similar to Fig. \ref{fig_expB31} this figure shows only the first 25000 points of the previous experiment shown in Fig. \ref{fig_lzw1}. 
\label{fig_lzw2}}
\end{figure}

With using the LZW dictionary tree to narrow the scope of selecting Q-actions, we can significantly improve Samu's learning as it is shown in Fig. \ref{fig_lzw1} and Fig. \ref{fig_lzw2}.
In our case, the nodes of the LZW tree are triplets. For example, the next log snippet shows the tree that has been built at the end of the first incremental step.  
\lstset { %
    language=C,
basicstyle=\ttfamily\footnotesize
}
\begin{lstlisting}
____2__ son was Hezron
__1__ sons were Zerah 
____2__ sons were Zerah
__1__ sons were and
__1__ son was Amminadab
______3__ son was Amminadab
____2__ son was Ram
__1__ son was Hezron
______________7__ son was Amminadab
____________6__ son was Ram
__________5__ son was Hezron
________4__ sons were Zerah 
______3__ sons were and
____2__ son was Jacob  
__1__ son was Isaac    
__1__ son was Jacob    
____2__ son was Amminadab
__1__ son was Ram
0__ 
\end{lstlisting}
where the number in the beginning of each line shows the depth of a given triplet. During our experiments, the maximum depth of the LZW tree is restricted to 10. The modified version of Alg. \ref{alg-sq} is shown in Alg. \ref{alg-sq2}.

\begin{algorithm}[h!]
\small
\caption{Q-learning-with-LZW-narrowing ($s'$, $t'$) \textbf{returns} a triplet (an action)\label{alg-sq2}}
\begin{algorithmic}[1]
\Require $s'$, $t'$ \Comment{$s'$ is the current state and $t'$ is the current triplet.}
	\Statex $r', a', p, nn, q, \mathcal{B}$ \Comment{Local variables: $r'$ is the current reward, $a'$ is the current action and $p$ is a perceptron, $nn, q \in \mathbb{R}$, $\mathcal{B}$ is a set of triplets.}
	\Statex $s, r=-\infty, a, N_{sa}, Prcps_{a}, \mathcal{N}$\Comment{Persistent variables: $s$ is the previous state, $r$ is the previous reward, $a$ is the previous action, $N_{sa}$ is the state-action frequency table and $Prcps_{a}$ is the array of the perceptrons, $\mathcal{N}$ denotes the actual node of the LZW tree.}
\Ensure $a'$\Comment{$a'$ is an action, the predicted next triplet.}
\Procedure{$\operatorname{SamuQNN}$}{$s'$, $t'$}
\State $r' = compare(t', a)$ 
\State $\mathcal{N} = build\_lzw\_tree(t')$ 
\State $a' = t'$ 
  \If{$r > -\infty$}
	\State Increment $N_{sa}[s,a]$
	\State $nn = Prcps_{a}[a](s)$
	\State $\mathcal{B} = \{\text{children of the node }  \mathcal{N}\}$	
	\State $q = nn +\alpha(N_{sa}[s,a]) (r' + \gamma\ max_{p \in \mathcal{B}} Prcps_{a}[p](s') - nn)$
	\State Back-propagation-learning($Prcps_{a}[a], s, q, nn$) \Comment{where $q-nn$ is the error of the MLP $Prcps_{a}[a]$ for input $s$.}
	\State $a' = argmax_{p \in \mathcal{B}} f(Prcps_{a}[p](s'), N_{sa}[s',p])$
\EndIf
\State $s = s'$
\State $r = r'$
\State $a = a'$
  
\State \Return{$a'$}
\EndProcedure
  \end{algorithmic}
\end{algorithm}

Finally, we may remark that Amminadab applies some cellular automaton steps in his visual imagery:
\lstset { %
    language=C++,
basicstyle=\ttfamily\footnotesize
}
\begin{lstlisting}
      for ( int i {1}; i<nrows-1; ++i )
        for ( int j {1}; j<ncols-1; ++j )
          console2[i][j] = console[i-1][j]+console[i][j-1]
                                     +console[i+1][j]+console[i][j+1];
\end{lstlisting}
In addition in this experiment the SARSA  \cite[pp. 844]{RussellNorvig} version of Q-learning is applied because in this case it is not necessary to use the restriction $p \in \mathcal{B}$ introduced in line 9 of Alg. \ref{alg-sq2}. The precise details can be found in source files \texttt{samu.hpp} and \texttt{ql.hpp} in the Amminadab's GitHub repository \cite{amminadabrepo}.

\section{Summary and highlights of Samu}

Samu B\'atfai is a disembodied developmental robotic chat software agent that is intended to become the basis of a chat system which will be able to talk and read in natural language like humans do.
Samu
\begin{itemize}
\item[--]
    directly implements the definition of machine consciousness (such as the definitions of conscious and intuitive computer programs) introduced in the paper \cite{cop}.
\item[--]
    corresponds the principles of Developmental Robotics \cite{DevRobPrinc}.
\item[--]
    applies Q learning with neural networks approximators
\item[--]
    uses multilayer perceptrons for approximation of the Q function, it is a deep learning \cite{NatureQ}, \cite{deepnn1}, \cite{deepnn0} feature.
\end{itemize}

\section{Conclusion}

Now that we have a working prototype we will be able to start deep Q-learning experiments to investigate Samu's behavior in big data corpora. It is certainly an important problem in this direction is the triplet extraction.
We believe that this can be handled well with the existing NLP tools. 
As we highlighted earlier, we did not focus on this issue moreover we want to keep it apart from the mainstream of Samu development. But since it will be a very important data mining component of Samu we are going to try to develop it in programming competitions. The first competition has already been announced for our university students. For details, see the author's university page \url{http://shrek.unideb.hu/~nbatfai/}. 
In a similar situation we have already used competitions \cite{oocwc} to catalyze our research programs but this area in question has its own challenges like the Turing test or the Loebner prize \cite{loebner}. We would like to participate in these challenges in the future.

Returning to the investigation of Samu's behavior in big data corpora, in applying Alg. \ref{alg-sq} there is a principal problem that it uses the state as an index of the frequency table at this moment. This is not a viable option for real applications. But on the one hand the frequency table may be dispensed and on the other hand the current triplet also be used as a simplified state in indexing the frequency table. 
The next critical part of Samu's code is the big number of possible actions. Working at the level of 2  years old children Samu should use roughly 2000 words.  If we calculate only 700 (S) $\times$ 300 (P) $\times$ 700 (O) triplets  we have $1.47*10^8$ combinations.  Let's remember that each combination corresponds to a perceptron and the memory footprint of a perceptron is about 40 megabytes of memory right now. 
But it is also true that this is the worst case, because it strongly depends on the settings of the perceptrons and it can be reduced drastically by using Jacob's character console-based visual imagination.
And fortunately, the distribution of these  combinations is Pareto rather than uniform so this aspect also seems to be handled well. 

As additional further work we plan to develop an intellectual (fractal) dimension \cite{ImitGame} based benchmark 
to test the intellectual capabilities of Samu type agent programs. In our case the intellectual dimension may be derived naturally from the triplet prediction used by Samu type agents.

In summary, in this paper we have begun to develop rapid prototypes to support experiments using the deep reinforcement learning techniques to facilitate the creation of a chat robot that will be able to talk in natural language like a human does. The generic name of these prototypes is Samu. 
At this moment Samu has already worked formally as chat program but despite that we thought of him as a child who is in the prenatal development stage. According to the plans Samu will be learning in family circle when he is born. We hope this paper can help you fork your own family chatbot from Samu projects.

\section*{Acknowledgment}

The computations used for the experiments shown in this paper were partially  performed on the NIIF High Performance Computing supercomputer at University of Debrecen. But it is important to note that the results in this paper can be reproduced on an ordinary PC.

\bibliography{samu}


\end{document}

%% file: samuarc1.tex
\begin{tikzpicture}[scale=0.92]

\node at (-3.75,4.5) {sentences};

\draw  (-2,5) rectangle (0.5,4);
\node at (-.75,4.5) {\begin{tabular}{c}
2D simula- \\ tion game
\end{tabular}};

\node (v1) at (-3,4.5) {};
\node (v2) at (-2,4.5) {};
\draw [-latex] (v1) edge (v2);

\draw  (1.5,5) rectangle (4,4);
\node at (2.75,4.5) {\begin{tabular}{c}
screen- \\ shots
\end{tabular}};

\node (v3) at (.5,4.5) {};
\node (v4) at (1.5,4.5) {};
\draw [-latex] (v3) edge (v4);

\draw  (5,5) rectangle (7.5,2.5);
\node (v5) at (4,4.5) {};
\node (v6) at (5,4.5) {};
\draw [-latex] (v5) edge (v6);

\node at (6.25,3.75) {\begin{tabular}{c}
Q-learning
\end{tabular}};

\draw[dashed]  (-0.5,3.5) rectangle (2.5,2.5);
\node (v7) at (2.5,3) {};
\node (v8) at (5,3) {};
\draw [-latex] (v7) edge (v8);

\node at (1,3) {\begin{tabular}{c}
reality \\ 
\end{tabular}};
\node at (3.75,3.25) {score};

\draw [-latex] (7.5,3.75) -- (8,3.75) -- (8,2) -- (-2.5,2) -- (-2.5,4.5);
\node at (3,1.75) {controls};

\end{tikzpicture}

%% file: samuarc2.tex
\begin{tikzpicture}[scale=0.92]

\draw  (-5.5,5) rectangle (-4.5,4);
\node at (-5,4.5) {NLP};

\draw  (-3.5,7) rectangle (-1.5,5.5);
\draw  (-3.5,7) rectangle (-2,6);
\node at (-3,6.5) {\begin{tabular}{c}
simul \\ prog
\end{tabular}};

\node at (-2.5,5.75) {running};

\draw [-latex] (-4.5,4.5) -- (-4,4.5) -- (-4,6.5) -- (-3.5,6.5);

\draw [-latex] (-4,5) -- (-4,2.5) -- (-3.5,2.5);

\node (v1) at (-6.25,4.5) {};
\node (v2) at (-5.5,4.5) {};
\draw [-latex] (v1) edge (v2);

\draw  (-3.5,3) rectangle (-1.5,2);
\node at (-2.5,2.5) {comp};

\node at (-3.5,4.5) {SPO};

\node (v3) at (-2.5,5.5) {};
\node (v4) at (-2.5,3) {};
\draw [-latex] (v3) edge (v4);

\node at (-2,3.5) {SPO};

\node at (-1.8,5) {\begin{tabular}{c}
next \\ sentence
\end{tabular}};

\draw  (0.5,7) rectangle (3,2);

\node (v5) at (-1.5,6.25) {};
\node (v6) at (0.5,6.25) {};
\draw [-latex] (v5) edge (v6);

\node (v7) at (-1.5,2.5) {};
\node (v8) at (0.5,2.5) {};
\draw [-latex] (v7) edge (v8);
\node at (1.7,4.5) {Q-learning};
\node at (-0.5,6.5) {screenshots};
\node at (-0.5,2.75) {score};
\draw [-latex] (3,4.5) -- (3.5,4.5) -- (3.5,7.5) -- (-3,7.5) -- (-3,7);
\node at (0,7.75) {prg writing controls};
\end{tikzpicture}

%% file: samuarc3.tex
\begin{tikzpicture}[scale=0.92]


\draw  (-5.5,5) rectangle (-4.5,4);
\node at (-5,4.5) {NLP};

\draw  (-3.5,7) rectangle (-1.5,5.5);

\node at (-2.5,6.5) {\begin{tabular}{l}
simul prog\\ S.P(O); list
\end{tabular}};

\draw [-latex] (-4.5,4.5) -- (-4,4.5) -- (-4,6.5) -- (-3.5,6.5);

\draw [-latex] (-4,5) -- (-4,2.5) -- (-3.5,2.5);

\node (v1) at (-6.25,4.5) {};
\node (v2) at (-5.5,4.5) {};
\draw [-latex] (v1) edge (v2);

\draw  (-3.5,3) rectangle (-1.5,2);
\node at (-2.5,2.5) {comp};

\node at (-3.5,4.5) {SPO};

\node at (-2,1.75) {SPO};

\draw  (0.5,7) rectangle (3,2);

\node (v5) at (-1.5,6.25) {};
\node (v6) at (0.5,6.25) {};
\draw [-latex] (v5) edge (v6);

\node (v7) at (-1.5,2.5) {};
\node (v8) at (0.5,2.5) {};
\draw [-latex] (v7) edge (v8);
\node at (1.7,4.5) {Q-learning};
\node at (-0.5,6.5) {image};
\node at (-0.5,2.75) {score};
\draw [-latex] (3,4.5) -- (3.5,4.5) node (v3) {} -- (3.5,1.5) -- (-2.5,1.5) -- (-2.5,2);
\node at (0.5,1) {action = next statement};

\node (v4) at (4,4.5) {};
\node (v5) at (3.25,4.5) {};

\draw [-latex] (v5) edge (v4);
\node at (4,5) {\begin{tabular}{l}
potential\\response
\end{tabular}};
\node at (4,4) {SPO};

\end{tikzpicture}